\documentclass[journal]{IEEEtran}

\usepackage{graphicx}
\usepackage{amsmath,amsfonts,amssymb}
\usepackage{setspace}
\usepackage{tocloft}
\usepackage{graphicx}
\usepackage[caption=false]{subfig}
\usepackage{caption}
\usepackage{comment}
\usepackage{diagbox}
\usepackage{makecell}
\usepackage{multirow}
\usepackage{tabularx}
\usepackage{adjustbox}
\usepackage{enumitem}  
\usepackage{cite}
\usepackage{xcolor}
\newcommand{\mytilde}{\raise.17ex\hbox{$\scriptstyle\mathtt{\sim}$}}

\cftpagenumbersoff{figure}
\cftpagenumbersoff{table} 
\graphicspath{{./Images/}}

\ifCLASSINFOpdf
\else
\fi
\begin{document}
	%
	\title{Anomaly Detection for Solder Joints Using $\beta$-VAE}
	%
	%
	%
	

	\author{Furkan~Ulger,~Seniha~Esen~Yuksel,  Atila~Yilmaz
	
	\thanks{F. Ulger, S.E. Yuksel and A. Yilmaz are all with the Department
	of Electrical and Electronics Engineering, Hacettepe University, Ankara, 06800, Turkey. e-mail: furkan.ulger@stu.ee.hacettepe.edu.tr,  eyuksel@ee.hacettepe.edu.tr,\\ ayilmaz@ee.hacettepe.edu.tr}}
	
	
	%
	%

	\markboth{}{}
	%



	\maketitle
	
	\begin{abstract}
		In the assembly process of printed circuit boards (PCB), most of the errors are caused by solder joints in Surface Mount Devices (SMD). In the literature, traditional feature extraction based methods require designing hand-crafted features and rely on the tiered RGB illumination to detect solder joint errors, whereas the supervised Convolutional Neural Network (CNN) based approaches require a lot of labelled abnormal samples (defective solder joints) to achieve high accuracy. To solve the optical inspection problem in unrestricted environments with no special lighting and without the existence of error-free reference boards, we propose a new beta-Variational Autoencoders (beta-VAE) architecture for anomaly detection that can work on both IC and non-IC components. We show that the proposed model learns disentangled representation of data, leading to more independent features and improved latent space representations.
		We compare the activation and gradient-based representations that are used to characterize anomalies; and observe the effect of different beta parameters on accuracy and on untwining the feature representations in beta-VAE. Finally, we
		show that anomalies on solder joints can be detected with high accuracy via a model trained on directly normal samples without designated hardware or feature engineering. 
	\end{abstract}
	

	%
	\IEEEpeerreviewmaketitle

	\section{Introduction}
	In the electronics manufacturing industry, SMD assembly lines produce printed circuit boards (PCB) with high density at a very fast rate, which makes them prone to errors. Detecting solder joint errors in the early stages of the assembly is critical to reduce rework cost in production and scrap rates. However, this task is difficult due to the highly varied appearance of PCBs, given that even the electronic components used for the same purpose differ enormously in appearance. Furthermore, different types of solder joint errors can occur such as solder bridging, excessive/insufficient solder, pseudo solder, solder skips etc. Additionally, quality requirements standardized by the Institute for Printed Circuits should be met for high-quality production. In order to fulfil this need, Automated Optical Inspection (AOI) devices are used in assembly lines. These devices achieve very high recognition rates but require dedicated hardware. There are surveys \cite{moganti1996automatic,janoczki2013automatic} that elaborate on the assembly process and the importance of solder joint inspection in the manufacturing industry.

	With the advancement in deep learning, CNN, a powerful algorithm for feature extraction, is used for the inspection of solder joints. On the other hand, despite achieving high accuracy, supervised CNN based approaches require a lot of labelled defective (abnormal) samples for training, but there is a lack of abnormal samples although plenty of error-free (normal) samples are available in many datasets used for solder joint inspection. Moreover, abnormalities may occur due to various reasons, and take on various shapes and sizes which might be hard to capture in a data collection. Yet, to our knowledge, there is no open dataset that has enough abnormal samples, possibly due to confidential reasons.  
	
	Therefore, in this paper, we propose to treat the solder joint defect detection as an anomaly detection problem, and resort to deep generative models, specifically to $\beta$-VAE \cite{Higgins2017betaVAELB}, which enable learning solely from normal samples.
	$\beta$-VAE is a variation of VAE \cite{kingma2013auto,rezende2014stochastic} that maximizes the probability of generating real data while balancing the likelihood and Kullback–Leibler (KL) divergence terms with an adjustable hyperparameter $\beta$.
	The $\beta$ term puts an emphasis on learning statistically independent latent factors which lead to disentangled representations where one single latent unit is sensitive to changes in a single generative factor. This disentanglement property of the $\beta$-VAE leads to more successful and more interpretable representations \cite{bengio2013representation}.
	

	Once the model describing the normal data is estimated via $\beta$-VAE, we generate an anomaly score for each test sample and apply a decision rule to detect anomalies. For generating the anomaly scores, we employ gradient-based representations as well as activation-based representations such as the reconstruction error.
	Finally, the model specifically learned to describe normal data is expected to yield a low anomaly score for normal data, and a high anomaly score for abnormal data. Hence, we detect these anomalies by comparing the resultant score against a threshold. With better-disentangled representations obtained using $\beta$-VAE, we show that anomalies on solder joints can be detected with high accuracy via a model trained directly on normal samples. Further, we compare this algorithm to two state-of-the-art approaches, namely the VAE and the Convolutional Autoencoder (CAE), and show the increase in detection rates.  
	
	The main contributions of this study are the following: 
	\begin{enumerate}[label=(\roman*)]
		\item  We attempt to solve a unique solder joint inspection problem in unrestricted domains, as such, it does not require specialized lighting, feature engineering, abnormal samples in training nor an error-free reference board.  
		
		\item  We design a $\beta$-VAE architecture tailored for this problem, analyze it with different anomaly scoring techniques, and investigate the effect of $\beta$ for solder joint inspection. Further, we show the benefits of using $\beta$-VAE and show how the latent space representations of normal and abnormal solders become visibly separated. 
		
		\item Our model can inspect both Integrated Circuit (IC) and non-IC component solder joints with a single architecture. To the best of our knowledge, there is no such study within the scope of anomaly detection.
		
	\end{enumerate}
	
	In the rest of this paper, we describe the related work in Sec. \ref{Sec:relWork}, present our proposed method in Sec. \ref{sec:Methods}, discuss our data set, parameter settings and experiments in Sec. \ref{sec:Experiment} and provide the results in Sec. \ref{sec:Results}.

	\section{Related Work} \label{Sec:relWork}
	
\begin{figure*}
		\begin{center}
			\begin{tabular}{c}
				\includegraphics[width=3.0cm]{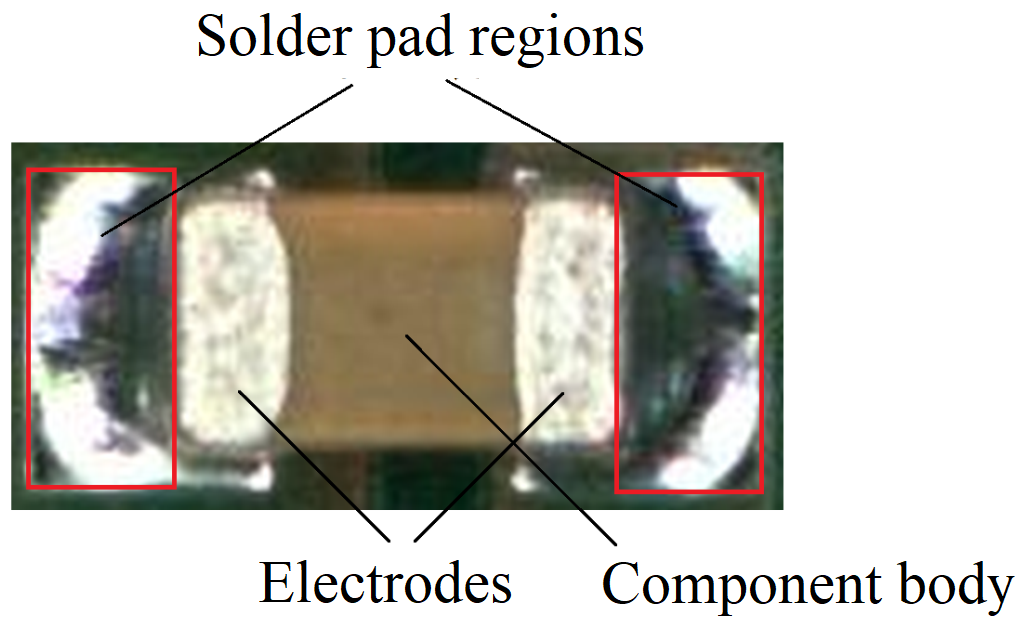} \hspace{1.5cm}
				\includegraphics[width=3.5cm]{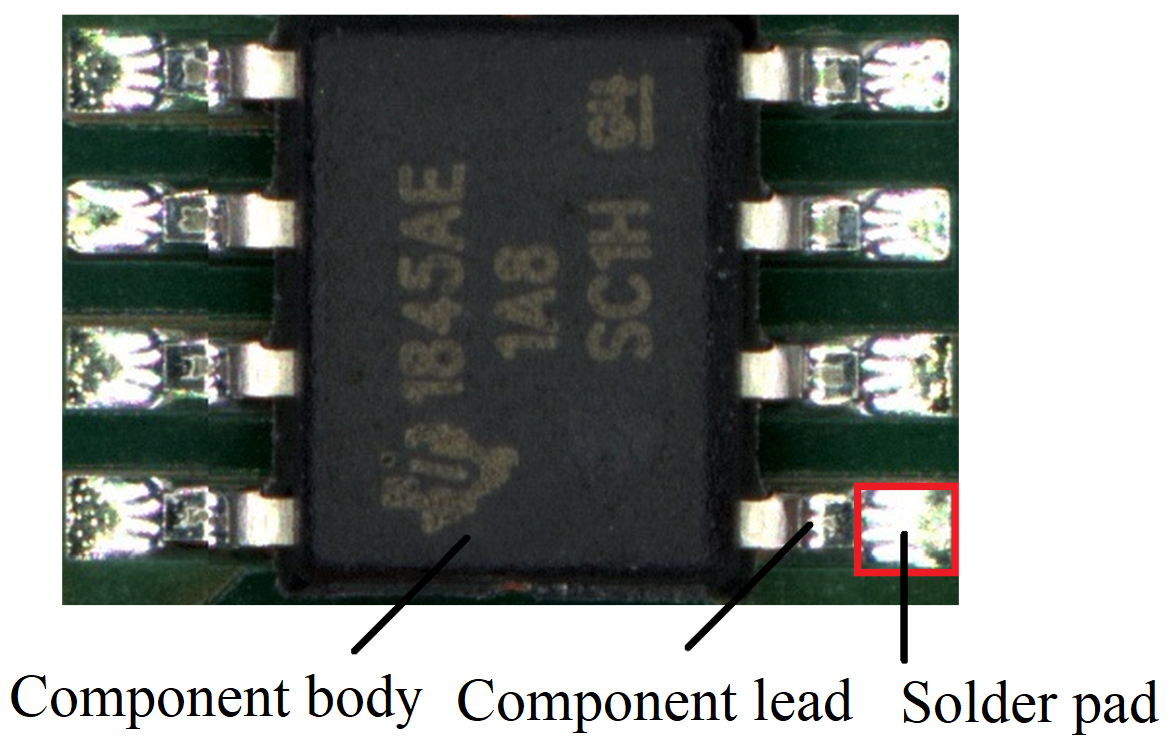}
				\\
				\hspace{-.5cm} (a) \hspace{4.25cm} (b)
			\end{tabular}
		\end{center}
		\caption 
		{
			Surface Mount Device (SMD) parcellation: (a) non-IC component image and its solder pad and component body, (b) IC component image and its body, lead and solder pad regions.} 
		\label{fig:nonIC_IC}
	\end{figure*}

	The literature for solder joint inspection can be divided into two main groups, namely, the methods based on feature extraction and the methods based on deep learning.
	Feature extraction based methods can mostly either detect IC or non-IC solder joint errors since there are no discriminative features presented for both types of components. Besides, their successful applications mostly rely on a  circular or hemispherical tiered RGB illumination for inspection of both non-IC component solder joints \cite{kim1999visual,hongwei2011solder,wu2013classification,wu2008aoi,wu2014inspection,song2019smt} 
	and IC component solder joints \cite{wu2011feature,ko2000solder}. On the other hand, there are some studies that do not require an RGB illuminator, but mostly an error-free reference image to inspect solder joints \cite{crispin2007automated, acciani2006application, ulger2019standalone, mar2011design}. The tiered illumination is used to generate colour contours on the solder paste, making use of the specular surface characteristics of the solder defects \cite{capson1988tiered}. These color contours are then analyzed via handcrafted features based on both color and shape to classify solder-joint errors. The most prominent color features are the average intensity and the percentage of highlight of each binarized RGB channel \cite{hongwei2011solder,  wu2013classification}. 
	Geometric (shape) features in frequent use can be listed as solder and polarity areas \cite{wu2008aoi}, barycenter and distribution features \cite{wu2014inspection} as well as the template matching feature obtained by normalized cross correlation (NCC) \cite{crispin2007automated}. Other proposed features include Wavelet coefficients \cite{acciani2006application} and Gabor features \cite{mar2011design}. Additionally, feature selection mechanisms are employed to reduce the number of features and to improve the classification performance \cite{crispin2007automated,  song2019smt}. 
	The classifiers employed in the more traditional literature include the Bayes classifier \cite{kim1999visual,wu2013classification}, multi-layer perceptron (MLP) \cite{ko2000solder, acciani2006application}, learning vector quantization (LVQ) \cite{wu2014inspection,ko2000solder} and Support Vector Machines (SVM) \cite{wu2013classification,dai2020soldering,song2019smt}; followed by the multi-stage classifiers \cite{kim1999visual,wu2013classification} to further improve accuracy. In practice, obtaining such reflections on a solder joint to extract these features is difficult; it requires special lighting and prior or manual input to parse the solder regions as shown in Fig. \ref{fig:nonIC_IC}. Additionally, traditional feature extraction based methods require determining distinctive features for each type of solder-joint errors.
	
	In recent years, there have been a few studies that employ deep learning based methods to inspect solder joints \cite{cai2018smt,dai2020soldering,mujeeb2019one}. Dai et al. \cite{dai2020soldering} employed the YOLO detector for solder localization and used semi-supervised learning for classification with SVM. Cai et al. \cite{cai2018smt} used cascaded CNNs: one learns the region of interest where solder defect is probable and the others are used to classify the solder joint of IC component. 
	
	These studies employed both anomalous and normal data, and  still used the tiered RGB illumination. Since anomalous data is rare, it is difficult to gather a representative training set for a supervised CNN to learn to identify all types of solder joint errors. Additionally, commonly inspected PCB defects in the literature are good solder, insufficient solder, excessive solder, pseudo solder, missing solder, missing component and tombstone; however there are more different types of possible PCB defects such as shifted and misplaced component, solder balling, solder flags, pin holes, solder discoloration, billboarding, dewetting and the like. Therefore, an alternative way to handle solder error detection would be the anomaly detection perspective. 
	
	For anomaly detection, Autoencoder \cite{zong2018deep} is a widely employed network that can learn to model complex data distributions. Autoencoders are trained directly on normal (error-free) data to minimize the reconstruction error such that, an encoding network, which strongly reduces the input to a compressed form, and a decoding network, which reconstructs its output to resemble the original input, are simultaneously trained. In the context of solder joint inspection, Mujeeb et al.\cite{mujeeb2019one} used a linear autoencoder on RGB illuminated solder joints and proposed using the $L2$ distance between the features of a reference image and a defective image as a similarity measure. However, the method requires a reference image that is a good (error-free) sample of the defective sample and tiered RGB illumination.

	Although Variational Autoencoders (VAE) are already being used for anomaly detection in different contexts such as dermatology \cite{lu2018anomaly} and medical imaging \cite{zimmerer2019unsupervised}; to the best of our knowledge, we are the first to use it for solder error detection and to offer a customized architecture for this problem. Also, our problem is unique, in that, we consider both IC and non-IC solder joints, we do not require any special lighting and neither do we require an error-free reference image. In doing so, we want to develop a more general solder-defect classifier that is not restricted to laboratory environments.

	\section{Methods} \label{sec:Methods}
	We briefly give an introduction to VAE and its variation, $\beta$-VAE. Latterly, scoring techniques that assign an anomaly score to each test data are presented.  
	\subsection{VAE and $\beta$-VAE}
\begin{figure*}[ht]
		\centering
		\includegraphics[width=5.5in]{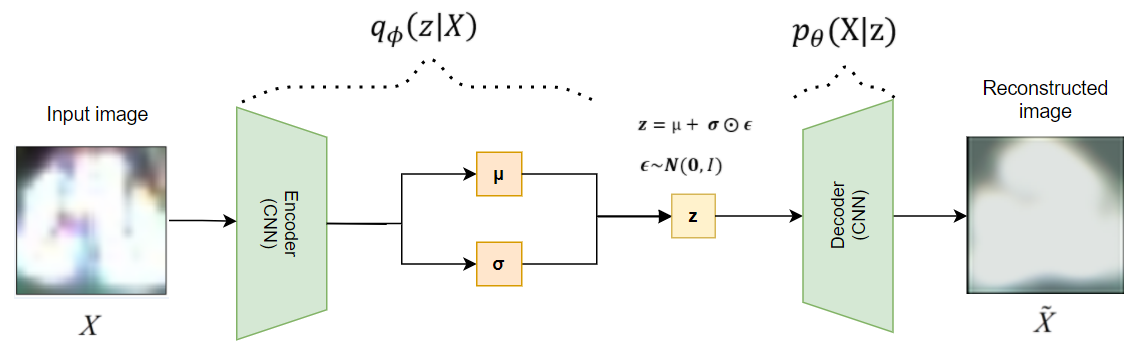}
		\caption{Variational Autoencoder Architecture with isotropic Gaussian prior distribution over latent vector.}
		\label{fig:VAEflow}
	\end{figure*}

	VAE is a generative model that maximizes the marginal likelihood 
	$p_{\theta}(\mathbf{X})$  where $\mathbf{X}$ is a data point (image) in a high dimensional space. It consists of a probabilistic encoder and a decoder where $\phi$ and $\theta$ are the encoder and decoder parameters respectively that include weights and biases, as shown in Fig. \ref{fig:VAEflow}. The encoder approximates the true posterior distribution $p_{\phi}(\mathbf{z}|\mathbf{X})$ whose distribution is assumed to be a Gaussian, where $\mathbf{z}$ is the latent (unobserved) vector of variables. The mean and standard deviation of the approximate posterior $q_{\phi}(\mathbf{z}|\mathbf{X})$ are the output of the encoder. 
	
	We also let the prior distribution have a standard Gaussian distribution $p(\mathbf{z})= N(0, \mathbf{I})$. Then, $\mathbf{z}$ is sampled from the posterior distribution as $\mathbf{z} \mytilde q_{\phi}(\mathbf{z}|\mathbf{X})= N(\boldsymbol{\mu}, \boldsymbol{\sigma})$.  The decoder is a generator that reconstructs $\tilde{\mathbf{X}}$ from $\mathbf{z}$.  However, we can not backpropagate the loss function to train the model since sampling operation is not differentiable. Therefore, a reparametrization trick is introduced. This trick uses a random variable $\boldsymbol{\epsilon}$ that is sampled from $N(0,I)$ and it is transformed to random variable z shifted by the mean and scaled by the standard deviation i.e. $\mathbf{z}= \boldsymbol{\mu} + \boldsymbol{\sigma} \odot \boldsymbol{\epsilon}$ that becomes a sample from the distribution $\mathbf{z} \mytilde N(\boldsymbol{\mu}, \boldsymbol{\sigma})$. 
	The illustration of VAE architecture is given in Fig. \ref{fig:VAEflow}. 
	
	To train VAE, the objective is finding the parameters $\theta$ and $\phi$ that maximize the evidence lower bound (ELBO) $\mathcal{L}$ of the marginal likelihood of a data point $p_{\theta}(\mathbf{X})$. The first term on the right-hand side of Eq. \ref{objective:VAE} is maximizing the probability of reconstructing the input $\mathbf{X}$ which corresponds to reconstruction loss. The second term is minimizing the KL divergence, which is a distance measure for probability distributions, between approximated posterior (encoder's distribution) and fixed Gaussian distribution with zero mean and unit variance. This leads to the lower bound on the probability density function of our data as is shown in Eq. \ref{objective:VAE}. We want to find the parameters (weights, biases) that maximize his objective function. Note that in the original VAE, $\beta=1$.
	
	\begin{figure*}
		\begin{equation}
			log p_{\theta}(\mathbf{X})\geq max_{\phi,\theta} \frac{1}{N} \sum_{n=1}^{N} log p_{\theta}(\mathbf{X_{n}} | \mathbf{z}) - \beta KL(q_{\phi}(\mathbf{z}|\mathbf{X}) || N(\mathbf{z}; 0, \mathbf{I}))
			\label{objective:VAE}
		\end{equation}
	\end{figure*}
	
	
	$\beta$-VAE introduces a weighting factor $\beta$ that balances reconstruction accuracy and latent channel capacity in Eq. \ref{objective:VAE}. With $\beta>1$ stronger pressure for the posterior $q_{\phi}(\mathbf{z}|\mathbf{X})$ to be close to the standard Gaussian prior $p(\mathbf{z})$ is introduced. This limits the capacity of latent $\mathbf{z}$ to learn a better disentanglement.  $\beta$-VAE with $\beta=1$ corresponds to the original VAE. Higgins et al. \cite{Higgins2017betaVAELB} demonstrated that $\beta$-VAE with the hyperparameter $\beta>1$ outperforms VAE on various datasets since it forces learning a more efficient latent representation of the data. This provides increased disentangling performance. However, there is a trade-off between reconstruction accuracy and learnt disentanglement quality within the learnt latent representations. A high $\beta$ value can lead to poorer reconstruction due to the loss of information while passing through the restricted latent channel. Therefore, it is important to find a balance that ensures a disentangled representation while still being able to reconstruct $\mathbf{X}$.
	On the other hand, $\beta$-VAE requires only tuning $\beta$ hyperparameter that can be estimated heuristically and does not suffer from the training instability of GAN-based approaches.
	

	\subsection{Anomaly Score}	
	In testing, samples $\mathbf{X}$ at the test set are inputted to the encoder of the trained VAE. The encoder outputs a vector of means and standard deviations of each image. Reparametrization trick is applied to get the latent vector $\mathbf{z}$ which is sampled from Gaussian distribution with the mean and standard deviation computed by the encoder. Then the decoder reconstructs the input image from this distribution. The first term in Eq. \ref{objective:VAE} forms the reconstruction loss $(\mathcal{A(\mathbf{X})}_{Recon})$. It is computed as the anomaly score by calculating the mean squared error between the input image and reconstructed image for each test sample, as given in Eq. \ref{sRecons} where N is the mini-batch size. VAE trained on only normal samples is expected to be able to reconstruct normal samples with low reconstruction error and yield low anomaly score, whereas abnormal samples are reconstructed with high reconstruction error and yield high anomaly score. 
	
	\begin{equation}
		\mathcal{A(\mathbf{X})}_{Recon}= log p(\mathbf{X}|\mathbf{z})=\frac{1}{N} \sum_{n=1}^{N} (\tilde{\mathbf{X}}_{n} - \mathbf{X_{n}})^2\\
		\label{sRecons}
	\end{equation}
	
	The objective is probability of occurrence of $\mathbf{X}$, hence anomaly score can also be calculated by $\mathcal{A(\mathbf{X})}_{ELBO}= -logp(\mathbf{X})$. Since the data is continuous and encoder's distribution is assumed to have a Gaussian distribution, KL divergence can be computed in closed form as the first term of the right hand-side of Eq. \ref{sELBO} as derived by Kingma and Welling \cite{kingma2013auto}: 
	
	\begin{figure*}
		\begin{equation}
			\mathcal{A(\mathbf{X})}_{ELBO}=  \frac{1}{2}\sum_{j=1}^{J}(1+log(\sigma^2_{j})-\mu_{j}^2-\sigma_{j}^2) - \frac{1}{N}\sum_{n=1}^{N} (\tilde{\mathbf{X_{n}}} - \mathbf{X_{n}})^2\\
			\label{sELBO}
		\end{equation}
	\end{figure*}
	
	Kwon et al. \cite{kwon2020backpropagated} proposed training autoencoders with gradient constraint to model normal data distributions and using gradient-based representations for anomaly detection, motivated to capture information unavailable in the activation-based network representation. Anomalies require more model parameter updates to be represented compared to normal data, hence backpropagated gradients are used to characterize how much model update is required by input to detect anomalies.
	The gradients of each decoder layer with respect to model parameters, $\frac{\partial\mathcal{L}}{\partial\theta_i}$ are calculated by backpropagating the reconstruction loss. The algorithm combines reconstruction and gradient loss as an anomaly score is given by, \[\mathcal{A(\mathbf{X})}_{GradCon}= \mathcal{A(\mathbf{X})}_{Recons} + \gamma \mathcal{L}_{grad} \] where $\gamma$ is a scalar and $\mathcal{L}_{grad}$ is cosine similarity i.e. the cosine of the angle between average training gradients and gradient loss of current ($k_{th}$) iteration as given in Eq. \ref{eq:gradCon}.
	
	\begin{figure*}
		\begin{equation}
			\mathcal{L}_{grad}= - \mathbb{E}_i\Bigg[cosSIM(\frac{\partial{J^{k-1}}}{\partial\theta_{i_{avg}}}, \frac{\partial \mathcal{L}^k} {\partial \theta_{i}})\Bigg], \hspace{0.5cm} 
			\frac{\partial{J^{k-1}}}{\partial\theta_{i_{avg}}}= \frac{1}{k-1} \sum_{t=1}^{k-1} \frac{\partial{J^t} }{\partial{\theta_i}}
			\label{eq:gradCon}
		\end{equation}
	\end{figure*}
	
	\noindent where $J$ is the loss function that is calculated during training and defined as, \[ J= \mathcal{L} + \alpha \mathcal{L}_{grad} \] where $\mathcal{L}$ is ELBO and $\alpha$ is a weight for the gradient loss. This anomaly detection algorithm using gradient constraint for training autoencoders and using a combination of reconstruction error and gradient loss as an anomaly score is called GradCon.
	
	
	\subsection{Decision Rule}
	After calculating the anomaly score, a decision rule is used to determine whether the sample is an anomaly or normal by comparing against a threshold \cite{Boracchi2020}. Anomaly score is expected to be low for the samples that the model is trained on i.e. the normal samples, whereas the model is not able to model the never seen abnormal samples which yield a high anomaly score. The samples with higher anomaly score than the threshold are evaluated as an anomaly and the ones below the threshold are normal samples. The threshold for the decision rule is considered as a hyperparameter and determined on the validation set. It is set to a value close to the anomaly score obtained on the validation set. 
	
	
	
	\section{Experiment}\label{sec:Experiment}
	
	In this section, we describe our dataset and the preprocessing methods we employ. Then, we provide comparisons between VAE, $\beta$-VAE, and the CAE. The CAE is an unsupervised model that learns a low-dimensional representation of the input \cite{hinton2006reducing}, selected due to its characteristic to separate normal from abnormal data.


	\subsection{Dataset and Preprocessing}
	Our dataset has $2513$ normal training, $567$ normal validation and $150$ test samples ($66$ abnormal and $84$ normal). Some abnormal samples are obtained from the open dataset \cite{lu2020fics}. The abnormal (defective solder joints) samples in the test set include solder bridge, insufficient solder, excessive solder, solder flags, solder dewetting, pin holes, flux residues, missing and shifted components. Ground truth labels of these samples are only provided during testing to evaluate anomaly detection performance. The dataset is obtained by segmenting both IC and non-IC solder joints from solder pad regions of several PCBs as shown in red in Fig. \ref{fig:nonIC_IC}. For data preprocessing, our data is cropped from the short side to preserve the integrity of solder joint images and resized to $64$x$64$. Then, the data is normalized to zero mean, unit variance to have the range from $-1$ to $1$, and horizontal flipping is applied as a random transformation to avoid overfitting but without increasing the dataset size.
	
	\subsection{Architecture and Hyperparameters}
	
\begin{figure*}[ht]
		\centering
		\includegraphics[width=5.2in]{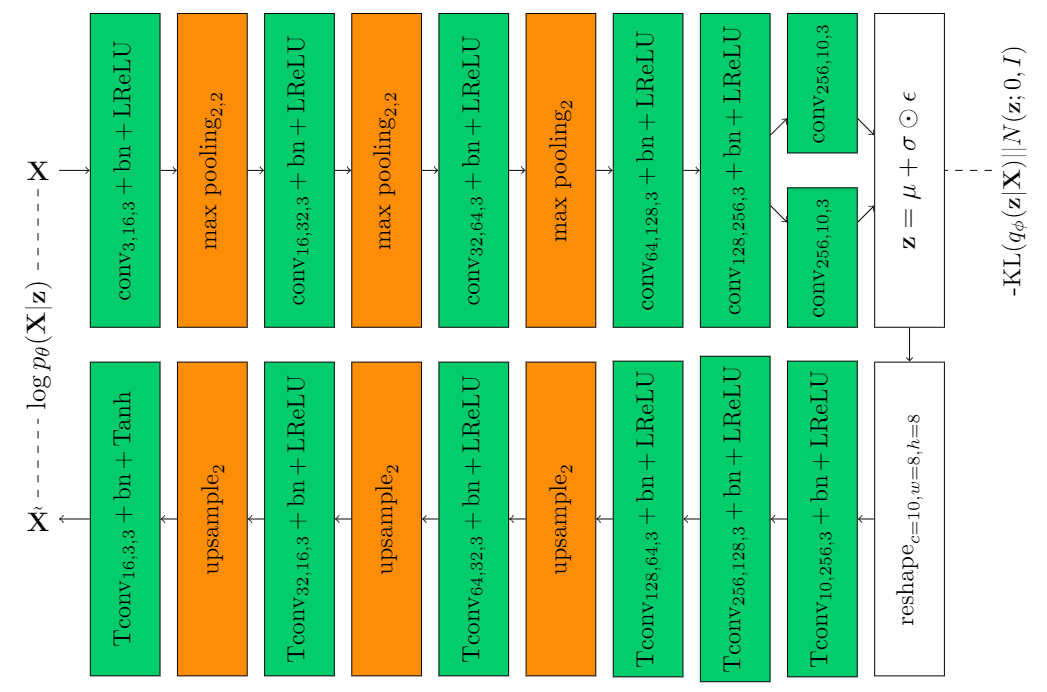}
		\caption{Sequence of blocks in the first row denote the encoder with $5$ convolutional (conv), $3$ max pooling layers and a convolutional bottleneck to obtain the parameters for the sampling of the latent vector $\mathbf{z}$. 
			Blocks in the second row form the decoder that uses $6$ transposed convolution (Tconv) and $3$ upsampling layers to reconstruct the input image from $\mathbf{z}$. The terms bn and LRELU stand for batch normalization and leaky RELU, respectively.}
		\label{fig:VAEarch}
	\end{figure*}

	Both VAE and $\beta$-VAE share the same architecture since the only difference is the added $\beta$ term in the objective function. These architectures were shown in Fig. \ref{fig:VAEflow} and in more detail with all the parameter sizes in Fig. \ref{fig:VAEarch}.  For comparison, the CAE has the same encoder-decoder structure except that there is a single convolutional layer at the bottleneck to compress the input. The encoder, which approximates the mean and standard deviation, has $5$ convolutional layers with an increasing number of filters from $3$ to $256$ and two separate convolutional layers at the bottleneck have $10$ filters. The decoder has transposed convolutional layers as a counterpart of the encoder. The convolutional layers have a filter size of $3$x$3$ and stride $1$. Convolutional layers are followed by batch normalization layers that normalize the feature map to zero mean and unit variance \cite{ioffe2015batch}. Leaky ReLU is applied for non-linearity to all the layers except for the output layer of the decoder, which uses Tanh. Max pooling layers are used to reduce the spatial dimension of feature maps by downsampling, while the upsampling layers increase the spatial size of the image by using bicubic interpolation. The VAE architecture is given in Fig. \ref{fig:VAEarch} with all the hyperparameters. As a convention, Conv$_{a,b,c}$ and Tconv$_{a,b,c}$ are the convolutional and the transposed convolutional layers, respectively, where $a$ is the number of channels in the input image, $b$ is the number of channels produced by the convolution, and $c$ is the size of the $c \times c $ convolution kernel. Max pooling$_{2,2}$ shows the max pooling filter size and stride, while upsample$_{2}$ provides bicubic interpolation sampling with a scale factor of $2$. The latent vector is reshaped to the channel size $10\times8\times8$ to generate the input image from latent vector $\mathbf{z}$.

	
	The model is trained with the ADAM optimizer \cite{kingma2017adam} for $100$ epochs with an initial learning rate of $1e-2$ that is decayed as the validation loss stops decreasing. To prevent overfitting to the training dataset, we resorted to $L2$ regularization that penalizes the high weights on features. The selected batch size is $64$. $\beta$ (KL weight) is selected as greater than one to learn better disentanglement of data, the effect of which will be more deeply analyzed in the next section. Lastly, the latent $\mathbf{z}$ dimension with the size of $640$ is selected since too small $\mathbf{z}$ can not model the marginal likelihood of a data point and too big $\mathbf{z}$ degrades the overall accuracy since learnt latent representation becomes indistinguishable for abnormal and normal samples. These hyperparameters given in Table \ref{hyperparameter1} are determined on the validation set.

	\begin{table}[ht]
		\caption{Hyperparameter Settings for $\beta$-VAE}
		\begin{center}
			\begin{tabular}{|c|c|}
				\hline
				\textbf{Hyperparameters}     & \textbf{Value} \\
				\hline
				Initial learning rate & $1e-2$   \\
				\hline
				Learning rate decay when loss plateous & $\gamma= 0.1$ \\ 
				\hline
				Number of Epochs     & $100$       \\
				\hline
				Optimizer     & Adam \\
				\hline
				Weight decay & $\lambda= 1e-4$ \\
				\hline
				Batch Size (N)     & $64$        \\
				\hline
				$\beta$	(KL weight)	   & $3$ \\
				\hline
				Latent space dimension & $640$ \\
				\hline
			\end{tabular}
			\label{hyperparameter1}
		\end{center}
	\end{table}
	
	\section{Results} \label{sec:Results}
\begin{figure*}[ht]
		\centering
		\includegraphics[width=5.8in]{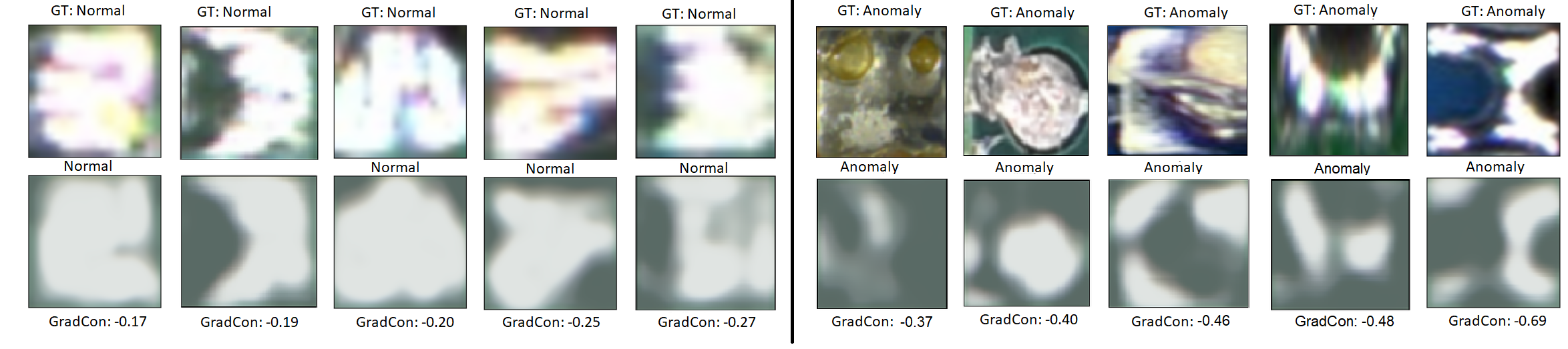}
		\caption{Reconstructed normal and abnormal samples. The first row shows original input images with corresponding ground-truth labels and the bottom row shows the reconstructions along with predicted labels and anomaly scores.}
		\label{fig:recons}
	\end{figure*}

	Original versus reconstructed normal and abnormal samples are given in Fig. \ref{fig:recons}. 
	Normal samples yield a low anomaly score, whereas abnormal samples yield a high anomaly score. Precision, recall and F1-score are used as evaluation metrics. Recall rate shows what proportion of actual anomalies are detected correctly and precision is the rate for the correctness of these estimations. F1-score is calculated to show the overall accuracy of the model from these metrics. All the results are at least an average of $50$ runs.
	

	The generative models, VAE and $\beta$-VAE, are compared based on the evaluation metrics given in Table \ref{methods}; namely the reconstruction loss (Recon), ELBO and the GradCon anomaly score (the combination of reconstruction and gradient loss). The CAE model was also evaluated with reconstruction loss and GradCon as an anomaly score. The highest accuracy is obtained for $\beta$-VAE  for $\beta= 3$ with a score of $0.809$ with GradCon anomaly, outperforming CAE and VAE. The ELBO loss resulted in lower accuracy than reconstruction loss for VAE and $\beta$-VAE. GradCon made improvement over all the models with activation-based representations.
	
	%
	
	\begin{table*}[ht]
		\caption{Comparison of reconstruction-based and generative models for different anomaly scores in terms of precision, recall and F1-score (average of 50 runs).}
		\begin{center}
			\begin{adjustbox}{width=\textwidth}
				\begin{tabular}{|c|c|c|c|c|c|c|c|c|} 
					\hline
					\multirow{2}{*}{\diagbox[innerwidth=2.7cm, height=1.0cm]{\textbf{Evaluation}}{\textbf{Method}}} & \multicolumn{2}{|c|}{\textbf{CAE}}  & \multicolumn{3}{|c|}{\textbf{VAE}} & \multicolumn{3}{|c|}{\textbf{$\beta$-VAE}}    \\ [0.5ex] 
					\cline{2-9}
					& Recon & GradCon & Recon & ELBO & GradCon & Recon & ELBO & GradCon    \\ [0.5ex] 
					\cline{2-9}
					\hline
					\multicolumn{1}{|c|}{Precision} & $0.728 \pm 0.01$ & $0.708$ & $0.779 \pm 0.05$  & $0.805 \pm 0.03$ & $0.760 \pm 0.03$ & $0.783 \pm 0.04$ & $0.825 \pm 0.03$ & $0.804 \pm 0.04$ \\
					\hline
					\multicolumn{1}{|c|}{Recall} & $0.573 \pm 0.01$ & $0.780 \pm 0.02$ & $0.802 \pm 0.03$ & $0.773 \pm 0.04$ & $0.836 \pm 0.04$ & $0.827 \pm 0.03$ & $0.777 \pm 0.04$ & $0.816 \pm 0.05$ \\
					\hline
					\multicolumn{1}{|c|}{F1-score} & $0.641$ & $0.742 \pm 0.01$ & $0.789 \pm 0.03$ & $0.787 \pm 0.03$ & $0.796 \pm 0.03$ & $ 0.803 \pm 0.03$ & $0.799 \pm 0.03$ & $\bf{0.809} \pm 0.04$\\  [1ex]
					
					\hline
				\end{tabular}
				\label{methods}
			\end{adjustbox}
		\end{center}
	\end{table*}
	
	Disentangled representations emerge when the right balance is found between the reconstruction accuracy and latent capacity restriction \cite{Higgins2017betaVAELB}, hence the effect of different $\beta$ term is investigated on $\beta$-VAE that is trained with gradient constraint as shown in Table \ref{beta}. $\beta=3$ is selected for $\beta$-VAE, which yields the highest F1-score. Using larger or smaller $\beta$ values also led to lower accuracy while larger beta values resulted in poorer reconstruction accuracy.
	
	\begin{table*}[ht]
		\caption{The effect of hyperparameter $\beta$ on accuracy for $\beta$-VAE trained with gradient constraint.}
		\centering
		\begin{tabular}{|c|c|c|c|c|c|} 
			\hline
			\diagbox[innerwidth= 2.9cm, height=1.2cm]{\textbf{Evaluation}}{\textbf{Beta} ($\beta$)} & \bf{0.01} & \bf{0.1} & \bf{1 (VAE)} & \bf{3} &  \bf{10} \\ [0.5ex] 
			\hline 
			Precision & $0.704 \pm 0.01$ & $0.734 \pm 0.01$ & $0.760 \pm 0.03$ & $0.804 \pm 0.04$ & $0.720 \pm 0.05$ \\
			\hline
			Recall  & $0.786 \pm 0.03$ & $0.824 \pm 0.02$ & $0.836 \pm 0.04$ & $0.816 \pm 0.05$ & $0.754 \pm 0.05$ \\
			\hline
			F1-score   & $0.743 \pm 0.02$ & $0.776 \pm 0.01$ & $0.796 \pm 0.03$ & $\bf{0.809} \pm 0.04$ & $0.735 \pm 0.04$ \\ [1ex]
			\hline
		\end{tabular}
		\label{beta}
	\end{table*}
	
\begin{figure*}
		\begin{center}
			\begin{tabular}{c}
				\includegraphics[height=3.0cm]{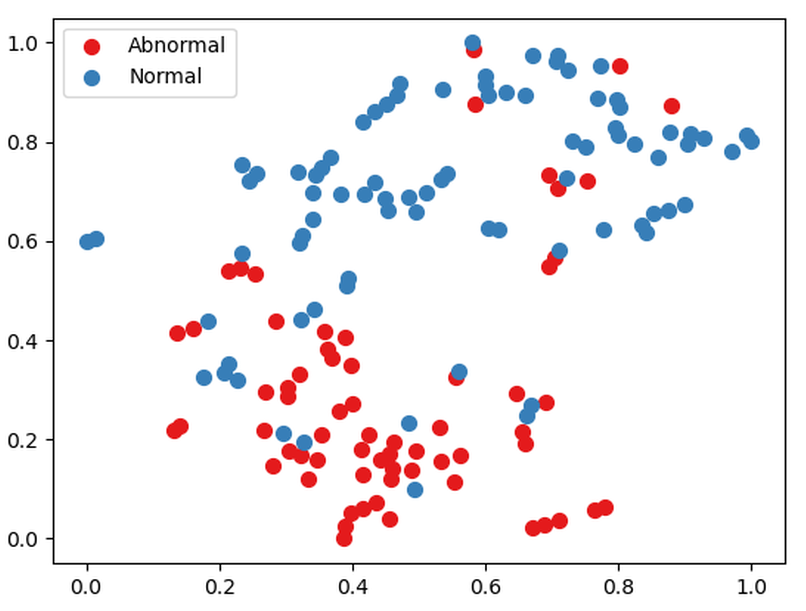}
				\includegraphics[height=3.0cm]{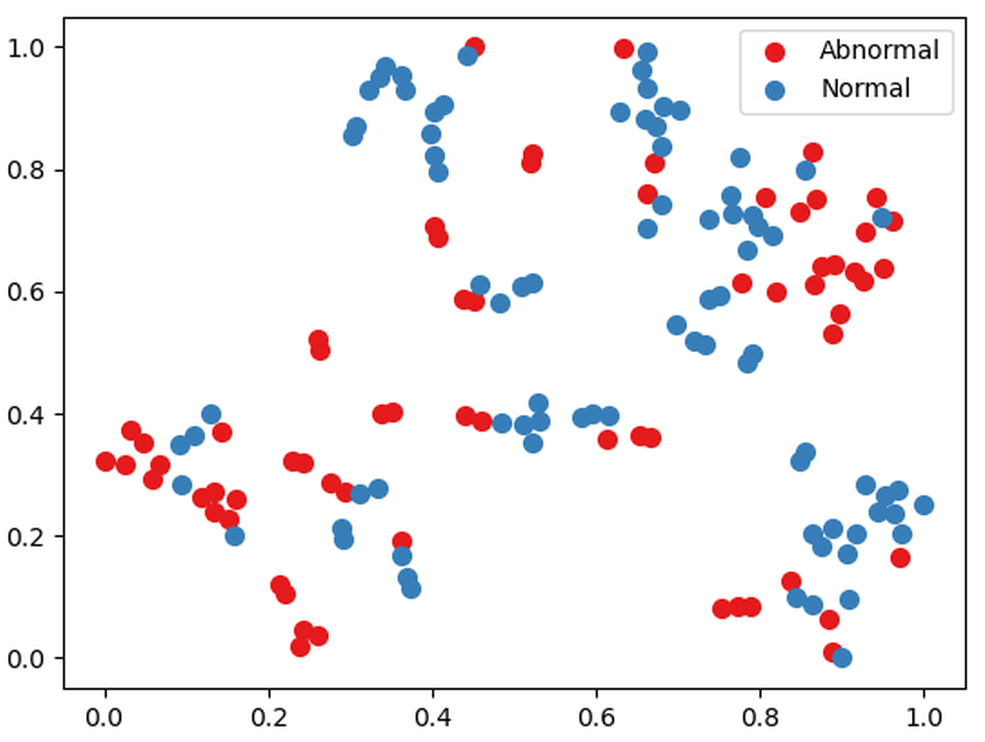}
				\includegraphics[height=3.0cm]{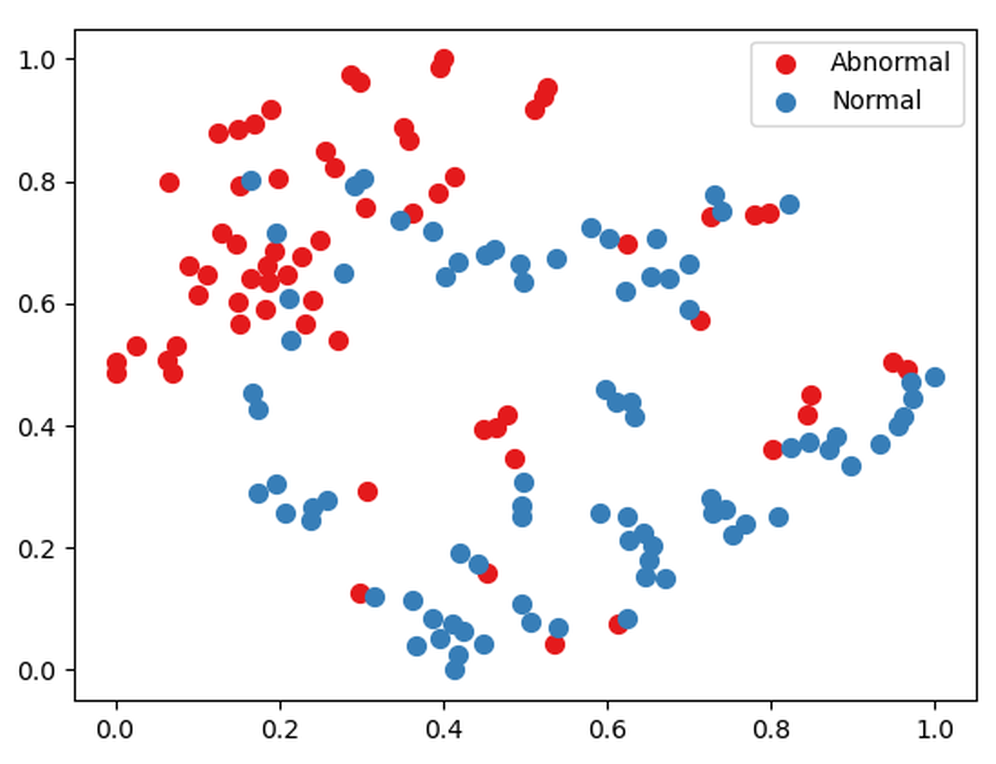}  
				\includegraphics[height=3.0cm]{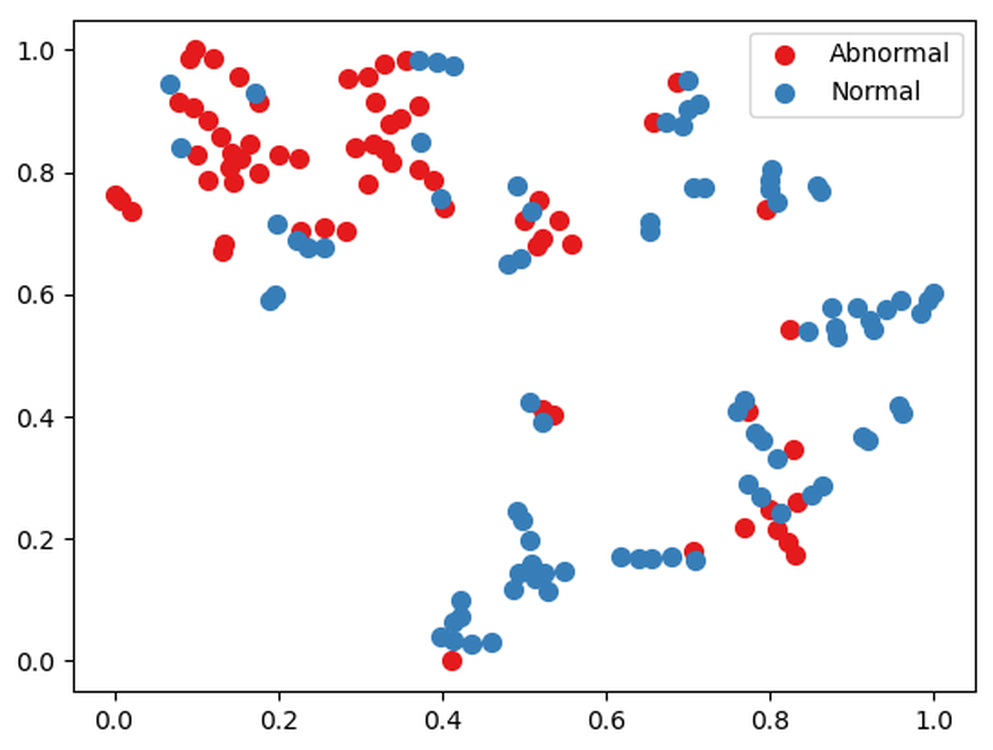}
				\\
				\hspace{-1.5cm} (a) \hspace{3.5cm} (b) \hspace{3.5cm} (c) \hspace{3.5cm} (d) \hspace{-1.5cm}
			\end{tabular}
		\end{center}
		\caption 
		{ \label{fig:tsne_Backprop}
			$2$-dimensional latent space representation of $\beta$-VAE that is trained with gradient constraint for (a) $\beta= 0.1$, (b) $\beta= 1$, (c) $\beta= 3$ and (d) $\beta= 10$ after applying t-SNE. Blue circles represent normal test data and the red circles represent abnormal test data.} 
	\end{figure*}

\begin{figure*}
		\begin{center}
			\begin{tabular}{c}
				\includegraphics[height=3.0cm]{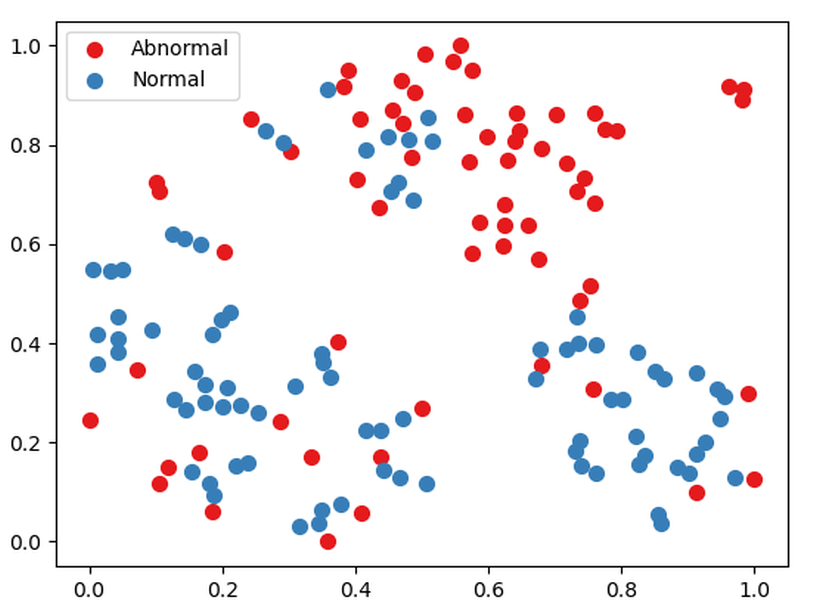}
				\includegraphics[height=3.0cm]{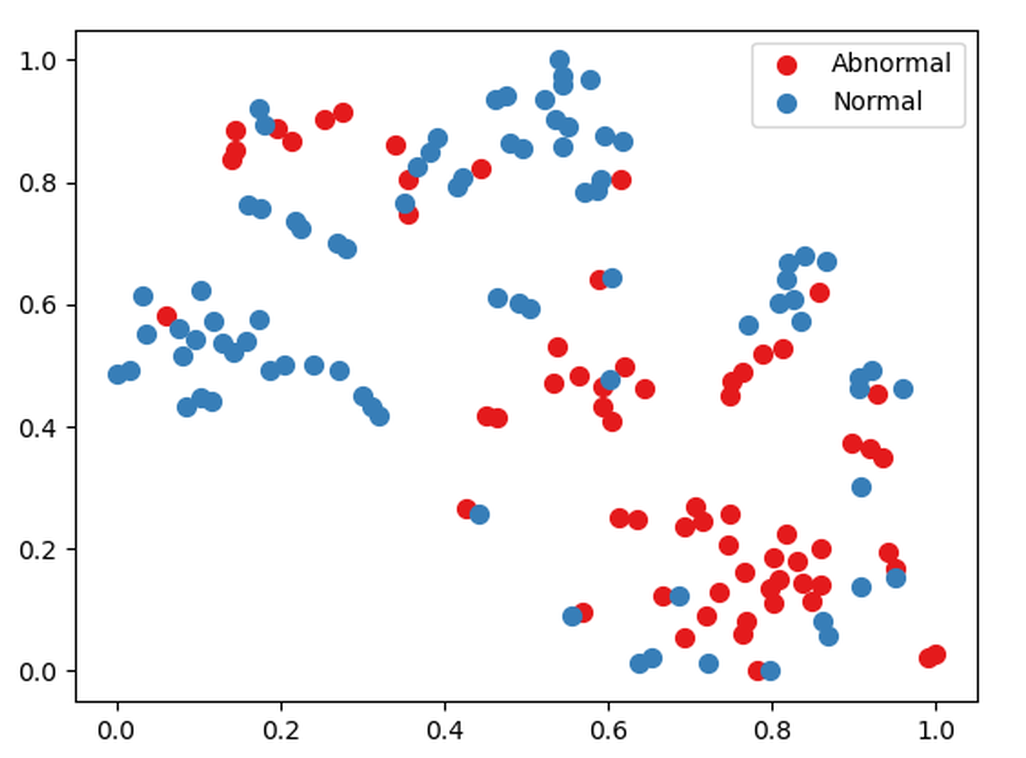}
				\includegraphics[height=3.0cm]{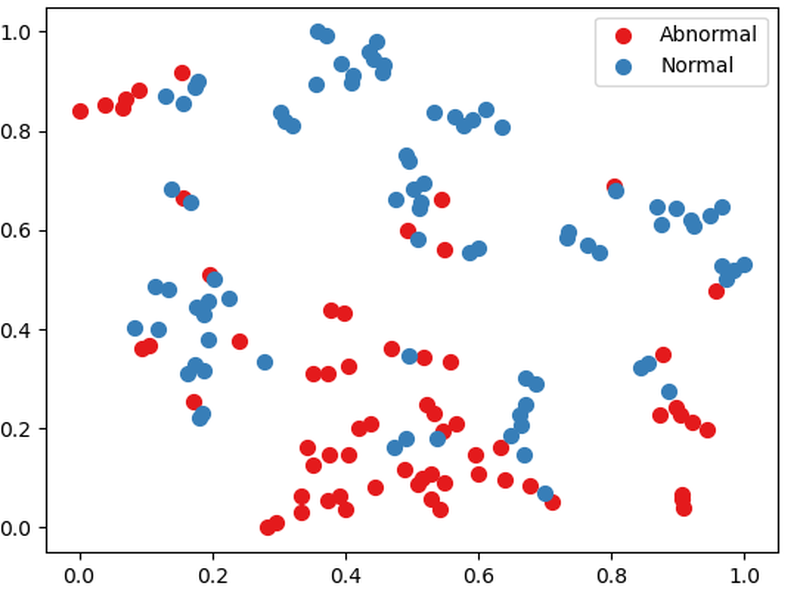}  
				\includegraphics[height=3.0cm]{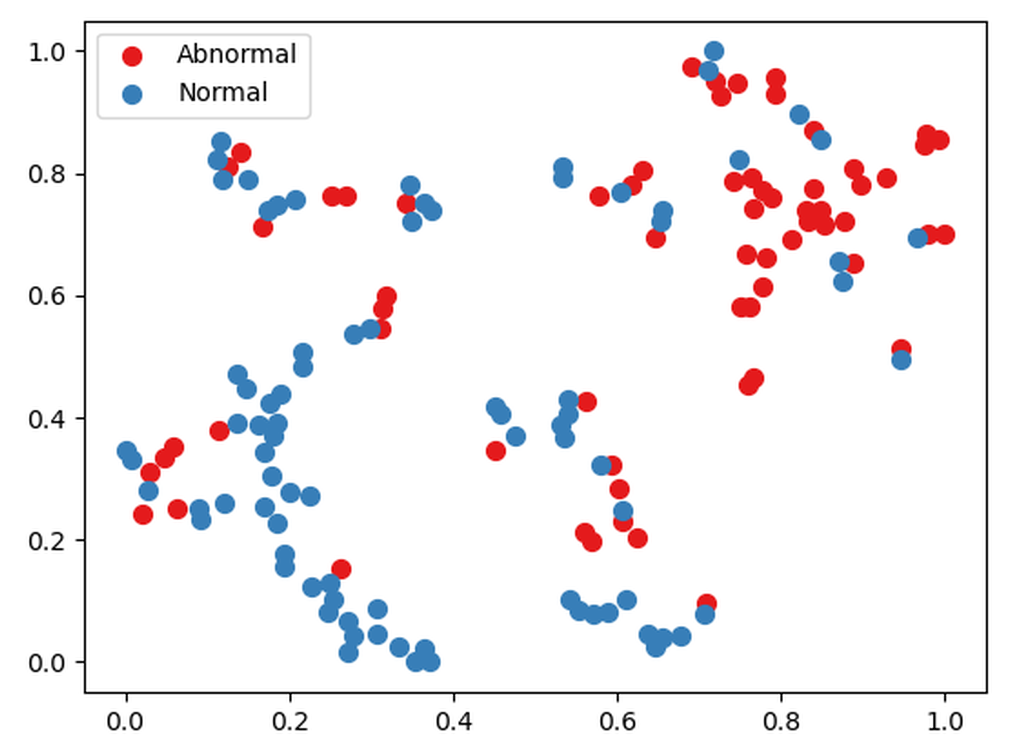}
				\\
				\hspace{-1.5cm} (a) \hspace{3.5cm} (b) \hspace{3.5cm} (c) \hspace{3.5cm} (d) \hspace{-1.5cm}
			\end{tabular}
		\end{center}
		\caption 
		{ \label{fig:tsne_ELBO}
			$2$-dimensional latent space representation of $\beta$-VAE that is trained with ELBO objective function for (a) $\beta= 0.1$, (b) $\beta= 1$, (c) $\beta= 3$ and (d) $\beta= 10$ after applying t-SNE. Blue circles represent normal test data and the red circles represent abnormal test data.} 
	\end{figure*}

	
	In order to see the effect of different $\beta$ values on the latent space representations, t-SNE \cite{van2008visualizing}, which is a nonlinear dimensionality reduction algorithm that is useful for visualizing high dimensional data, is employed. Test images are encoded in the beta-VAE that is trained with gradient constraint (GradCon) and ELBO objective function for different $\beta$ values, then the resultant mean vector of the approximated posterior distribution is passed to t-SNE with the perplexity of $5$ since the dataset is small. Then t-SNE maps $640$-dimensional latent space into $2$-dimensional space. The embeddings are scaled to bring all the values to $[0,1]$. Latent space visualization of $\beta$-VAE that is trained with gradient constraint and ELBO objective is given in Fig. \ref{fig:tsne_Backprop} and \ref{fig:tsne_ELBO} respectively. t-SNE is run $100$ times and the embeddings with the lowest KL divergence is selected for each model. As shown in Fig. \ref{fig:tsne_Backprop} and \ref{fig:tsne_ELBO}, samples from the same class are closer to each other and appear in clusters. For $\beta$-VAE that is trained with gradient constraint, clusters of the classes are entangled i.e. not easily separable for the $\beta$ values of $1$ and $10$ as shown in Fig. \ref{fig:tsne_Backprop} (b) and (d). This behaviour is observed for the $\beta$ values of $0.1$ and $10$ for $\beta$-VAE trained with ELBO objective as given in Fig. \ref{fig:tsne_ELBO} (a) and (d). However, we can see that the classes are more separable for $\beta= 3$ although they are split into parts as given in Fig. \ref{fig:tsne_Backprop} and Fig. \ref{fig:tsne_ELBO} (c) for both models. 
	For the real-world complex samples, it is difficult to achieve clear disentanglement or visualize it in two dimensions, but we can deduce that a better-disentangled representation is learnt for $\beta= 3$. 
	
	
	\section{Conclusions}
	In this study, solder joint defect detection in PCBs is considered as an anomaly detection problem. We propose using $\beta$-VAE, which is a generative model for disentangled representation learning, for anomaly detection to detect errors in both IC and non-IC component solder joints. beta-VAE disentangles factors of variation in data with that disentangled representations understanding and classifying data become easier. We compared different anomaly scoring techniques based on activation and gradient-based representations and the effect of different parameter beta on accuracy is investigated. The highest accuracy is achieved for the combination of reconstruction and gradient loss as an anomaly score to anomaly detection of solder joints.
	Despite the challenging environment of the problem due to the varying shape and small size of solder joints, we show that high accuracy can be achieved without using special lighting, hand-crafted features or labelled data for supervised learning. 
	In the context of solder joint inspection where normal samples are in abundance and there are few abnormal samples, we show that the designed beta-VAE architecture is effective and leads to disentangled latent space representations. 

	
	%

	



	\ifCLASSOPTIONcaptionsoff
	\newpage
	\fi

	
	
	\bibliographystyle{IEEEtran}
	\bibliography{Anomaly_Detection_for_Solder_Joints_Using_betaVAE.bib}
\end{document}